\documentclass[nohyperref]{article}

\usepackage{microtype}
\usepackage{graphicx}
\usepackage{subfigure}
\usepackage{booktabs} 

\usepackage{hyperref}

\usepackage{amsmath}
\usepackage{amssymb}
\usepackage{mathtools}
\usepackage{amsthm}

\usepackage[capitalize,noabbrev]{cleveref}

\usepackage{todonotes}

\newlength\myindent
\usepackage[accepted]{icml2022}

\icmltitlerunning{Model Based Meta Learning of Critics for Policy Gradients}

\begin{document}

\twocolumn[
\icmltitle{Model Based Meta Learning of Critics for Policy Gradients}



\icmlsetsymbol{equal}{*}

\begin{icmlauthorlist}
\icmlauthor{Sarah Bechtle}{mpi}
\icmlauthor{Ludovic Righetti}{nyu,mpi}
\icmlauthor{Franziska Meier}{fair}
\end{icmlauthorlist}

\icmlaffiliation{mpi}{Max Planck Institute for Intelligent Systems, Tuebingen, Germany}
\icmlaffiliation{nyu}{New York University, New York, USA}
\icmlaffiliation{fair}{META AI Research, Menlo Park, USA}

\icmlcorrespondingauthor{Sarah Bechtle}{sbechtle@tuebingen.mpg.de}


\icmlkeywords{Machine Learning, ICML}

\vskip 0.3in
]



\printAffiliationsAndNotice{}  
\begin{abstract}
Being able to seamlessly generalize across different tasks is fundamental for robots to act in our world. However, learning representations that generalize quickly to new scenarios is still an open research problem in reinforcement learning. In this paper we present a framework to meta-learn the critic for gradient-based policy learning. Concretely, we propose a model-based bi-level optimization algorithm that updates the critics parameters such that the policy that is learned with the updated critic gets closer to solving the meta-training tasks. We illustrate that our algorithm leads to learned critics that resemble the ground truth $Q$ function for a given task. Finally, after meta-training, the learned critic can be used to learn new policies for new unseen task and environment settings via model-free policy gradient optimization, without requiring a model.  We present results that show the generalization capabilities of our learned critic to new tasks and dynamics when used to learn a new policy in a new scenario. 
\end{abstract}
\section{Introduction}
\label{introduction}
\begin{figure}[ht]
\centering
    \includegraphics[width=0.45\textwidth]{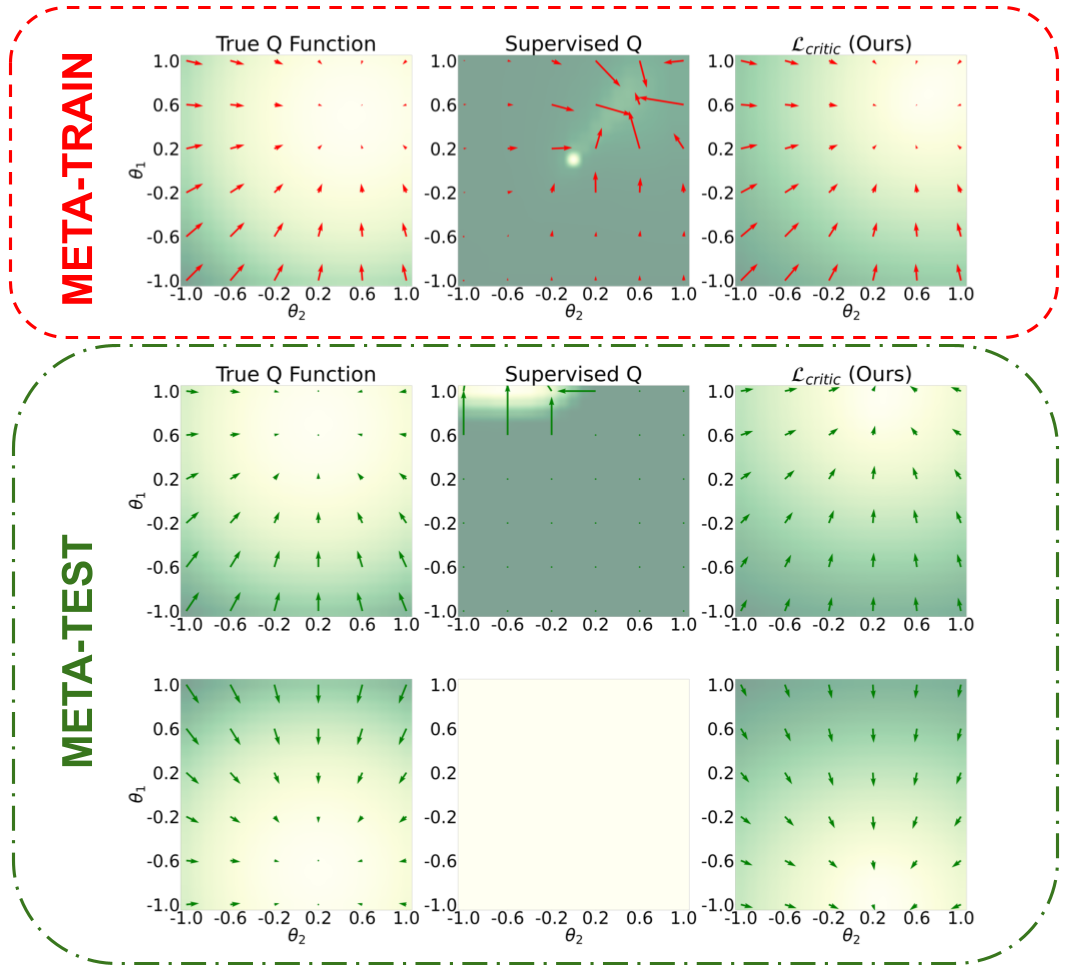}
    \caption{\small{
    Visualizing the policy optimization landscape induced by the true $Q_\text{true}$, the supervised $Q_\text{sup}$ function and our $\mathcal{L}_\text{critic}$ as a function of policy parameters $[\theta_1,\theta_2]=[x,y]$ (on the x,y axes). After meta training our $\mathcal{L}_\text{critic}$ resembles the gradient landscape of the true $Q$ function. At test time, our $\mathcal{L}_\text{critic}$ can be used to learn new policies for new goals from scratch since it generalizes the same way the true $Q$ function would. The color represents the computed state action value for each method when using the policy $\pi_{[\theta_1,\theta_2]}$ to collect trajectories.}}
    \label{fig:point_mass_teaser}
    \vspace{-15pt}
\end{figure}
Humans effortlessly generalize to new tasks and scenarios. Once we learned how to place a plate into the cupboard, we can quickly generalize this skill to placing a cup. This ability to generalize is a key aspect of human intelligence and is also important in reinforcement learning (RL):  we want the robot to be able to use what it previously learned to perform new tasks in new environments. In most RL learning settings however, the policy is tested in the same environment, under the same conditions it was trained on. Because of these shortcomings, the problem of generalization in RL is gaining popularity in the literature \cite{kirk2021survey}. The majority of work focuses on exposing the RL agent during train time to many different scenarios: different environments, different tasks or different random seeds. The goal is often to learn a policy or a latent representation that generalizes "as is" during test time to unseen conditions and tasks \cite{zhang2018study,zhang2018dissection,kirk2021survey}. Some work focuses on meta-learning more abstract representations such as a loss function \cite{bechtle2021meta} that can transfer to learn new policies at test time.

We believe that learning the learning signal (critic) for the policy is the representation needed to allow generalization and quick adaptation to new tasks and scenarios and thus follow this line of work. Standard approaches \cite{lillicrap2015continuous,mnih2015human} learn $Q$ functions (critics) in a supervised fashion, by using data collected through rollouts of the policy. Alternatively, approaches to meta-learn the critic have been proposed \cite{zhou2020online,sung2017learning}. In contrast to these approaches, our framework meta-learns the critic $\mathcal{L}_\text{critic}$ following a model based bi-level optimization procedure. 
%
We show that our meta-learned critic resembles a ground truth $Q$ function and compared to standard Q-function learning approaches, exhibits superior generalization performance when transferred to new tasks. Specifically, our $\mathcal{L}_\text{critic}$ can be used to learn new policies for new tasks under new dynamics.  We illustrate this with a point mass task, where we train our $\mathcal{L}_\text{critic}$ on the task of reaching a goal. In Figure \ref{fig:point_mass_teaser} (top row) 
we compare the optimization landscape of the true $Q_\text{true}$, the $Q_\text{sup}$ function learned in a supervised way and of our $\mathcal{L}_\text{critic}$ for the training task. Concretely, we plot the gradient fields for $\nabla_{[\theta_1,\theta_2]}Q_\text{true}$,$\nabla_{[\theta_1,\theta_2]}Q_\text{sup}$ and $\nabla_{[\theta_1,\theta_2]}\mathcal{L}_\text{critic}$ as a function of policy parameters $\mathbf{\boldsymbol{\theta}}=[\theta_1,\theta_2]$ 
We test the obtained Q functions (critics) on new goals which have new corresponding optimal policies $\mathbf{\boldsymbol{\theta}}^*_\text{test}$ (bottom rows). Our meta-learned critic $\mathcal{L}_\text{critic}$ resembles the gradient field of $Q_\text{true}$ after training and generalizes better to new unseen goals - as $Q_\text{true}$ would - since it trains policies with optimal parametrization for the new tasks. Further details are presented in section \ref{sec:what_learning}. 

To summarize, the contributions of this work are: (i) we present a framework to meta-learn the critic for continuous control policies (ii) we show how to use the model of the robot dynamics as physical inductive prior during meta-training while keeping our framework model-independent at meta-test time. This makes our learned critic also applicable in situation when the model changes at test time. Which is particularly relevant in the robotic domain, where analytical models of the robots are known and can be used, but it is difficult to update them every time the robot wants to place, for example, a new object into the cupboard. 
(iii) we provide an analytical and illustrative analysis on a simple point mass task that sheds light on what and how our critics are learned and generalize. We show that our meta learned critic framework at convergence recovers the optimal policy and generalizes at test time to new goals and policy initialization. (iv) We finally perform a quantitative evaluation on high-dimensional continuous control tasks. In short, we find that our meta-learned critic not only solves the RL problem of accomplishing the desired task during meta-train time, 
but also allows for generalization across multiple unseen tasks, new dynamics and policy initialization at test-time.
\section{Related Work and Background} \label{sec:related_work}
We start by reviewing traditional Q-function learning approaches, and then discuss work investigating generalization in RL and more concretely meta-learning for RL.
\paragraph{Problem setting}\label{sec:rl_setting}
We consider a Markov Decision Process consisting of states $s \in S$, actions $a \in A$, reward function $r(s,a) \in R$ and transition function $T(s_{t+1},a_t,s_t)$, that defines the transition to state $s_{t+1}$ when in state $s_t$ and applying action $a_t$. The goal of reinforcement learning is to learn a policy $a_t \leftarrow \pi(s_t)$, that predicts an action given a state and maximizes the expected return. A critic, generally speaking, provides a loss $\mathcal{L}$ for the policy $\pi$ that trains the policy to maximise the expected return. In this work, we learn the critic $\mathcal{L}_\text{critic}$ and use it to directly compute gradients $\nabla_\pi \mathcal{L}_\text{critic}$ to train the policy.
\paragraph{Supervised Learning of a Critic}\label{sec:sup_learning_q}
Neural fitted Q iteration, first presented in \cite{riedmiller2005neural}, aims to implement the classic Q-learning rule,
\begin{equation}\label{eq:Q_function}
    Q(s_t,a_t) = r_t + \gamma \max_a Q(s_{t+1},a_{t+1})
\end{equation}
where $\gamma$ is a discount factor that specifies the importance of long term rewards \cite{sutton2018reinforcement}, in a neural network. For this, $Q(s,a)_\phi$ is represented as a neural network, parametrized by $\phi$. An error function
\begin{align}\label{eq:q_error_function_sup}
   \delta = \sum_t (Q(s_t,a_t) - (r_t + \gamma Q(s_{t+1},a_{t+1})))^2
\end{align}
is defined to learn parameters $\phi$ in a supervised way. The $Q$ function is trained on batches of triples $(s_t,a_t,s_{t+1})$ collected using policy $\pi$ in the environment. \cite{mnih2015human} show in DQN how a policy $\pi$ for discrete action spaces is  learned to maximize the learned action value function $\max_{\pi(s_t)}Q(s_t,\pi(s_t))$. In \cite{silver2014deterministic} and \cite{lillicrap2015continuous} a policy $\pi_\theta$ is learned for continuous action spaces, using the neural fitted $Q$ function by taking a deterministic policy gradient update $\nabla_\theta Q(s_t,\pi_\theta(s_t))$ to find the  parameters $\theta^*$ of $\pi$ that solve the task. This allows for learning a continuous control policy, which is also the objective of this work. In \cite{schaul2015universal} the authors present an extension to DQN, to the multi-goal setting. In this case the Q function is learned over multiple goals $g \in G$, and the goal becomes an input to the neural network representing the Q function $Q(s_t,a_t,g)$. Also \cite{andrychowicz2017hindsight} is concerned with the multi-goal setting. In this work stored transitions $(s_t,a_t,s_{t+1})$ are relabeled depending on the current goal $g$. A general overview of learning a $Q$ function in a supervised way is given by algorithm \ref{algo:sup_q}. Most of the related work focuses on the single goal setting and all of the related work evaluates their approaches on single tasks settings, where training and testing happens on the same setting.  We believe a benefit of learning critics should be to re-use them to train new policies on new tasks. In other words, a critic should be a representation that generalizes and thus allows for more generalization in RL.  Our algorithm meta-learns a critic by directly evaluating the performance of the policy under the critic. This allows for generalization to new settings at test time.
\begin{algorithm}
\begin{algorithmic}[1]
\STATE{$\phi, \gets$ randomly initialize parameters of $Q_\phi$}
\STATE{$\theta \gets$ randomly initialize parameters of $\pi$}
\WHILE{not done}
\STATE{$\{s_t,a_t, r_t\}_{t=0}^T \gets \text{ rollout policy } \pi_{\theta}$}
\STATE{add $\{s_t,a_t, r_t\}_{t=0}^T$ to Dataset $D$}
\STATE{\text{\emph{Update $\phi$ using $D$:}}}
\STATE{$\phi \gets \phi - \eta \nabla_\phi \delta \text{ (Eq. \ref{eq:q_error_function_sup})}$}
\STATE{\text{\emph{Update $\theta$ using learned $Q$:}}}
\STATE{$\theta \gets \theta -\alpha \nabla_\theta \sum_{t}  Q(s_t,\pi_\theta(s_t))$}
\ENDWHILE
\end{algorithmic}
\caption{Supervised learning of $Q$ function}
\label{algo:sup_q}
\end{algorithm}

\paragraph{Generalization in Reinforcement Learning} 
As \cite{kirk2021survey} points out in a recent survey, most of RL algorithms work in a setting where the policy is evaluated on exactly the same environment it was trained on. This assumption would not hold in real-world scenarios and is also in contrast to the classical supervised learning setting, where training and testing sets are disjoint. The problem of generalization in RL is also studied in \cite{zhang2018dissection, zhang2018study} 
and both works find that RL is prone to memorization and overfitting, 
In \cite{cobbe2019quantifying} the authors observe that a large number of training environments are needed for generalization, in addition they also show how different neural network architectures can improve generalization. \cite{zhao2019investigating} in particular analyse the effects of domain shifts, like a changing mass on a robot end-effector, between training and testing and observe that current RL algorithms are particular vulnerable to this setting. Current work on generalization for RL focuses on studying better policy representations that better transfer to new scenarios, here we propose a meta-learning framework that learns a critic that can be transferred to learn new policies for new tasks and dynamics. 
\paragraph{Meta-Learning for RL}
One goal of learning to learn~\cite{Schmidhuber:87long, bengio:synaptic, ThrunP98} - also known as meta learning, it to learn representations that generalize to new tasks \cite{hospedales2020meta}. 
For example \cite{maml, mendonca2019guided, gupta2018meta,yu2018one} meta-learn model parameters for adaptation to new tasks.\cite{maclaurin2015gradient, l2l, li2016learning, franceschi2017forward, meier2018online, rl2} learn parameters update rules with respect to known loss or reward functions as a more general representation of the learning problem. Other work \cite{bechtle2021meta,sung2017learning, epg18, zou2019reward} learns loss/reward function representations. Recently, the concept of learning a loss function has also been applied to the RL domain in the form of critic learning \cite{zhou2020online,sung2017learning,xu2020meta}. Our framework falls into the category of learning a critic for RL. Similar to works by \citet{sung2017learning} and \citet{zhou2020online}, we aim at learning a critic that can then be used to optimize a new policy. In contrast to previous work, we learn our critic following a model based bi-level optimization procedure where the critic is learned by following the increase in performance of the new policy, given the model. In other words, directly following the learning progress of the policy. 
%
%
\section{Model-Based Meta-Learning of Critics}
Next, we present our approach of meta-learning the critic $\mathcal{L}_\text{critic}=M(s_t,a_t,g)$ for policy gradients. We are interested in the standard reinforcement learning problems defined in section \ref{sec:rl_setting}. $\mathcal{L}_\text{critic}$ is used to find the optimal parameters $\theta^*$ of policy $\pi(\theta)$ with gradient descent by computing a policy gradient $\nabla_\theta \mathcal{L}_\text{critic}$. In contrast to other related work that learn objective functions \cite{zhou2020online,lillicrap2015continuous}, we propose a model based algorithm that learns $\mathcal{L}_\text{critic}$ by leveraging a differentiable transition function $T$ during meta-train time. We show that $\mathcal{L}_\text{critic}$ is learned in a sample efficient way, generalizes to new tasks and dynamics and resembles the properties of the state action value function $Q(s_t,a_t)$ defined by Equation \ref{eq:Q_function}.
 \begin{figure}[H]
     \centering
     \includegraphics[width=0.3\textwidth]{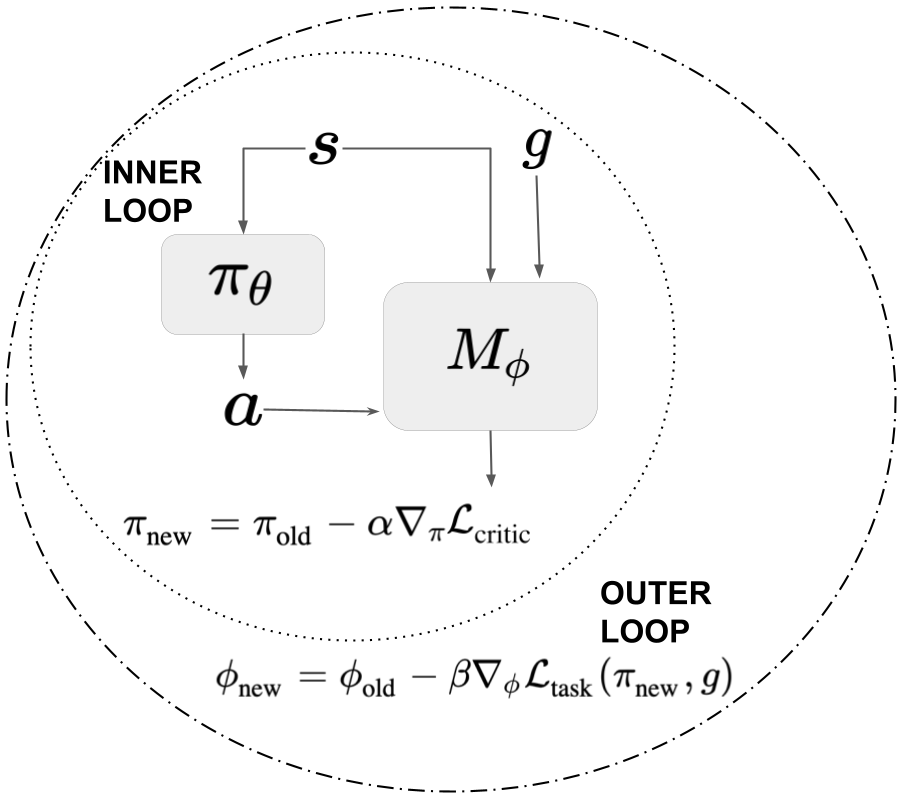}
     \caption{Our bi-level meta learning framework, to learn the critic we differentiate through the inner loop gradient update.}
     \label{fig:bi-level-meta}
     \vspace{-10pt}
 \end{figure}
\subsection{Overview: Bi-Level Meta-Learning of Critics}\label{sec:algorithm}
Our algorithm follows the bi-level loss learning framework presented in \cite{bechtle2021meta}.  Figure \ref{fig:bi-level-meta} shows an overview of the bi-level optimization procedure.

\textbf{The inner loop} uses $\mathcal{L}_\text{critic}$ to train the policy $\pi$, via policy gradient updates, where $\alpha$ is the learning rate:
\begin{equation}\label{eq:update_theta}
    \theta_\text{new}(\phi) = \theta_\text{old} - \alpha \nabla_\theta \mathcal{L}_\text{critic}.
\end{equation}
Notably, the final $\theta_\text{new}$ is a function of critic parameters $\phi$. 

\textbf{The outer loop} learns parameters $\phi$ of a goal-conditioned critic $\mathcal{L}_{\text{critic},\phi}(s_t,a_t,g)$. More concretely, a task loss $\mathcal{L}_\text{task}(\pi_{\theta_\text{new}},g)$ is used to update $\phi$ via gradient descent:
\begin{equation}
    \phi_\text{new} = \phi_\text{old} - \beta \nabla_\phi \mathcal{L}_\text{task}(\pi_{\theta_\text{new}(\phi)},g)
\end{equation}
where $\beta$ is the outer loop learning rate.  Intuitively, the task loss evaluates the performance of the updated policy $\pi(\theta_\text{new})$ in accomplishing task $g$. It is implicitly a function of $\phi$, since $\theta_\text{new}$ is a function of $\phi$. 
%
Given the task loss our framework differentiates through the inner loop to update $\phi$. It thus directly measures the learning progress of $\pi$ when using $\mathcal{L}_\text{critic}$ - which is in contrast to other meta learning approaches for reinforcement learning \cite{zhou2020online,sung2017learning}. Designing, the task loss, such that it is differentiable with respect to the policy parameters $\theta$ is one of the main challenges. In the next section we show how to achieve this by using a model.

\subsection{Model Based Meta Learning of Critic}
\begin{algorithm}
\begin{algorithmic}[1]
\STATE{$\phi, \gets$ randomly initialize parameters of $\mathcal{L}$}
\WHILE{not done}
\FOR{ $g \in G$}
\STATE{$\theta \gets$ randomly initialize policy parameters}
\STATE{\text{\textbf{INNER LOOP}}}
\STATE{$\{s,a, R\}_{t=0}^T \gets \text{forward unroll } \pi_{\theta}$}
\STATE{$\pi_{\theta_{\text{new}}} \gets \theta_\text{old} - \alpha \nabla_\theta \sum_{t=0}^T \mathcal{L}_\text{critic}(s_t,a_t,g)$}
\STATE{\text{\textbf{OUTER LOOP}}}
\STATE{$\{s_\text{new},a_\text{new}, R_\text{new}\}_{t=0}^T \gets \text{unroll } \pi_{\theta_\text{new}} \text{ using } f$}
\STATE{\text{\emph{Update $\phi$ to maximize reward under $f$:}}}
 \STATE{$\phi \gets \phi - \eta \nabla_\phi \sum_t \mathcal{L}_\text{task}(s_\text{new},a_\text{new}, R_\text{new})$}
 \ENDFOR
\ENDWHILE
\end{algorithmic}
\caption{Model Based Meta Learning}
\label{algo:ml3_mbrl}
\end{algorithm}
Our Model Based Meta Learning approach borrows the key idea of model based reinforcement learning (MBRL): to use a model during training. In contrast to MBRL however, our framework does not use the model to learn the policy directly and thus does not need the model during test time. In MBRL, a differentiable model of the dynamics of the robot $f$ is used to learn or optimize the policy $\pi$ \cite{atkeson1997comparison}. $f$ can be learned from experienced data $\mathcal{D}$ in an end to end fashion \cite{deisenroth2011pilco,chua2018deep,bechtle2020curious}. Alternatively analytical differentiable models of the robot dynamics or kinematics \cite{sutanto2020encoding} as well as differentiable simulators \cite{heiden2021neuralsim,heiden2019real2sim} can be used, enabling direct gradient based optimization of the policy. Independently of the modelling choice for $f$, the goal of model based reinforcement learning is to find a policy maximizing the expected reward $R$, given the model $\max_{\pi} \mathbb{E}_{f}[R_\pi]$. 

Instead of performing policy updates, our Model Based Meta learning framework uses a model $f$ to compute the outer loop task loss $\mathcal{L}_\text{task}$ during meta-training.: 
\begin{align}\label{eq:cumulative_reward}
       \mathcal{L}_\text{task}(\pi_{\theta_\text{new}(\phi)},g) &=  R_{\pi_{\theta_\text{new}}}^f
        = \sum_{t=0}^T r(f(s_t,\pi_{\theta_\text{new}}(s_t))). 
\end{align}
As long as $f$ is differentiable, the task loss is now differentiable with respect to $\pi$'s parameters $\theta_\text{new}$. We can leverage this property to update $\phi$ since $\theta_\text{new} = \theta_\text{old} - \alpha\nabla_\theta\mathcal{L}_\text{critic}(\phi)$:
\begin{equation}\label{eq:learned_loss_param_update}
    \phi_\text{new} = \phi_\text{old} - \beta \nabla_\phi (\mathcal{L}_\text{task}(\pi_{\theta_\text{new}(\phi)},g)).
\end{equation}
In words, the optimization learns a critic maximizing the expected cumulative reward under the new policy $\pi_{\theta_\text{new}}$ given model $f$. Algorithm \ref{algo:ml3_mbrl} provides an overview of the method.

To learn $\pi$ in the inner loop, we use the learned critic $\mathcal{L}_\text{critic}$
\begin{equation}\label{eq:mfrl_objective}
    \pi_{\theta_\text{new}} = \pi_{\theta_\text{old}} - \alpha \nabla_{\pi}\mathcal{L}_\text{critic}\nabla_\theta \pi.
\end{equation}
The above equation is a policy gradient objective that can be used to learn a policy in a model free way. This means that in the inner loop (as well as during meta-test time) $\mathcal{L}_\text{critic}$ is used to directly learn a policy for a new task without requiring the model $f$. In this work, however, we want to show how analytical differentiable models of the robot can be used as strong physical priors during meta learning.
We show in Section \ref{sec:what_learning} that our learned critic $\mathcal{L}_\text{critic}$ resembles in shape a the $Q$ \cite{sutton2018reinforcement} function after training. 
\subsection{Using the critic $\mathcal{L}_\text{critic}$ at test time}
After meta-learning, $\mathcal{L}_\text{critic}$ is used without requiring the model (see Algorithm~\ref{algo:ml3_mbrl_test}). This makes our framework model-free at test time. In section \ref{sec:experiments} we show how, even in cases where the dynamics $f$ change, our  $\mathcal{L}_\text{critic}$ is able to learn a policy that solves the new task without requiring the changed model. Our framework thus allows to exploit strong physical priors during meta-training but is flexible to changes in the dynamics by becoming model-free at test time.
\begin{algorithm}
\begin{algorithmic}[1]
\footnotesize{
\STATE{$\theta \gets$ randomly initialize policy; sample $g$}
\WHILE{not done}
\STATE{$\{s,a, R\}_{t=0}^T \gets \text{rollout } \pi_\theta $}
\STATE{$\pi_{\theta_{\text{new}}} \gets \theta_\text{old} - \alpha \nabla_\theta \sum_{t}\mathcal{L}_\text{critic}(s_t,a_t,g)$}
\ENDWHILE
}
\end{algorithmic}
\caption{Using $\mathcal{L}_\text{critic}$ to learn $\pi$ for new tasks.}
\label{algo:ml3_mbrl_test}
\end{algorithm}
\section{Analysis of learned critic $\mathcal{L}_\text{critic}$}
In this section, we further analyse our $\mathcal{L}_\text{critic}$.
During meta-train time we train on full trajectories that are collected from policy roll-outs. For the task loss $\mathcal{L}_\text{task}$ we consider the cost along the full trajectory (cf. Equation \ref{eq:cumulative_reward}). We update parameters $\phi$ of $\mathcal{L}_\text{critic}$ as shown in Equation \ref{eq:learned_loss_param_update}. 
We get the gradient update
\begin{equation} \label{eq:learned_loss_update}
    (s_T-g)\nabla_\phi \theta \nabla_\theta \pi \nabla_\pi \Gamma_{s_{T\dots0}}
\end{equation}
where $\Gamma_{s_{T\dots0}}$ is the trajectory from $s_{T}$ to the initial state $s_0$ and $s_T = f(f(f(\dots f(s_0,\pi(s_0))),\pi(s_{T-1}))$. Looking at Equation \ref{eq:learned_loss_update}, we see that the learned critic $\mathcal{L}_\text{critic}$ is updated by measuring the effect that changes in the policy parameters have on the observed trajectory. Since $\mathcal{L}_\text{critic}$ is used to learn $\pi$, the policy parameters change as a function of $\phi$. Thus the goal is to update $\phi$ to minimize the error of $\pi$ in performing the desired behaviour $g$. This means that $\mathcal{L}_\text{critic}$ is learned while trying to converge to the optimal policy $\pi(\theta^*)$. Thus, at convergence of the bi-level optimization loop: $\mathcal{L}_\text{critic}(s_t,a_t) \rightarrow \mathcal{L}_\text{critic}(s_t,\pi^*(s_t))$.

This is different from other approaches of $Q$ function learning. Instead of fitting the $Q$ function in a supervised way as in \cite{mnih2015human,lillicrap2015continuous}, our learned $\mathcal{L}_\text{critic}$ is learned in a self-supervised way by trying to increase the learning progress of $\pi$. Next we provide an analysis to show the benefits of learning a critic with our method. 
\subsection{Analysis: Recovering optimal policy}
With the help of a simple example, we analyze how our updates to the critic's parameters are significantly different when compared to standard supervised q-fitting. Furthermore we show how our method, at convergence of inner and outer loop, recovers the optimal policy, while standard q-fitting does not. This example has been proposed in \cite{fairbank2008reinforcement} where it has been shown that value learning methods \cite{sutton2018reinforcement} would not necessarily find the optimal policy. 
We define the following learned critic $\mathcal{L_\text{critic}}$, model $f$, policy $\pi$, reward $r$ and learned $Q$ function:
\begin{equation}
    \begin{split}
        \mathcal{L_\text{critic}}(s_t,a_t) = Q(s_,a_t) =  = (s_t + a_t)^2\phi_1\\
        f(s_t,a_t) = s_t+a_t = s_{t+1}\\
        a_t = \pi_\theta(s_t) = \theta \\
        r_t(s_t,a_t) =\begin{cases}
      0 & \text{if }t<T \\
      (s_t-g)^2 & \text{if }t=T\\
    \end{cases}    
    \end{split}
\end{equation}
Where $g$ is the goal. We consider a time horizon of $T=2$, following the example above at $t=2$ the goal will be $g = s_0 + 2\theta$ thus the optimal policy to reach a desired goal should be $\pi(s) = \theta = \frac{g - s_0}{2} $. 
The goal of this simple example is to i) show how different the update strategies are between our algorithm and standard q-fitting and ii) compare the achieved optima of the learning loops. In the following we assume we have access to a roll-out $\tau = \{s_0,a_0, r_0, s_1, a_1, r_1, s_2, r_2\}$.
\subsubsection{Supervised $Q$ learning}\label{sec:analysis_simple_q}
Let's first consider the case where we learn $Q$ from data, as presented in section \ref{sec:sup_learning_q}. 
Given $\tau$ we first compute the optimal parameters of $Q$ then use that Q to optimize the policy to convergence. 
We start with the Q function $Q_\phi(s_t, a_t) = (s_t + a_t)^2\phi_1$.
Using the rollout $\tau$ we compute the supervised loss signal $\delta$ for $Q_\phi$ according to Equation \ref{eq:q_error_function_sup}:
\begin{align}
\delta = \sum_{t=0}^{1} (Q_\phi(s_t,a_t) - (r_t + \gamma Q_\phi(s_{t+1},a_{t+1})))^2
\end{align}
Then we take the gradient of $\delta$ wrt. $\phi_1$ set the gradient to $0$ and solve for $\phi_1$, which results in optimal $\phi$:
\begin{align}
    \phi_1 &= \frac{(s_1+a)^2r_2}{[((s_0+a)^2-(s_1+a)^2)^2 + (s_1+a)^4]} \\
\end{align}
We have now identified the optimal $\phi$ as a function of the rollout data. More details on this step can be found in the Appendix \ref{sec:illapp_q_fitting}. Next we compute the gradient for policy parameters $\theta$ with respect to the predictions of the learned $Q_\phi$ function: 
\begin{align}
    &\nabla_\theta \sum_{t=0}^{T=1} Q(s_t, a_t)\\
    &=  \nabla_\theta ((s_0 + \theta)^2 \phi_1 + (s_1+\theta)^2 \phi_1 + r_2)\\
    &= 2\phi_1[(s_0+\theta)+(s_1+\theta)]
\end{align}
where we have added a terminal reward $r_2$ and have expanded the actions in the Q function predictions with the policy parametrization. This allows us to compute the analytical gradients wrt to $\theta$, set the gradient to $0$ ($\nabla_\theta \sum_{t=0}^{T=1} Q_\phi(s_t, a_t) = 2\phi_1[(s_0+\theta)+(s_1+\theta)=0$) and solve to find the optimum. This will converge to (see Eq. \ref{app_eq:convergence_q} in \ref{sec:illapp_q_fitting} for details):
\begin{align}
    \theta = \frac{s_0+s_1}{-2}
\end{align}
As we can see this does not recover the optimal policy.
We can perform the same analysis using a model $f$ during learning and will reach a similar result (see Appendix \ref{sec:ill_meta_critic}).
The meta critic learning framework presented in \cite{zhou2020online} also learns the critic in a supervised way an thus has a similar result, details are presented in Appendix \ref{sec:ill_meta_critic}. 
\subsubsection{Model Based Meta-Learning the critic}
Next we analyze the learning procedure of our method. We first update the parameters $\theta$ using our $\mathcal{L}_\text{critic}$ (in the inner loop). To do so, we compute $\nabla_\theta \sum_{t=0}^{T=1}\mathcal{L}_\text{critic} = \nabla_\theta [(s_0+\theta)^2\phi_1 + (s_1+\theta)^2\phi_1] = 2\phi_1(s_0+s_1+2\theta) \Rightarrow \theta_\text{new} = \theta- \alpha 2\phi_1(s_0+s_1+2\theta)$. We choose $\alpha=\frac{1}{2} \Rightarrow \theta_\text{new} = \theta-\phi_1(s_0+s_1+2\theta)$. As in Algorithm \ref{algo:ml3_mbrl} we use $\theta_{new}$ to compute $\mathcal{L}_\text{task}$ defined per Equation \ref{eq:cumulative_reward}:
\begin{equation}
    \mathcal{L}_\text{task} = \sum_t r_t = r_0 + r_1 + r_2 = r_2 = (s_2-g)^2
\end{equation}
expanding all terms, meaning $s_2 = s_1+\theta_\text{new}$, $s_1=s_0+\theta_\text{new}$, $\theta_\text{new} = \theta-\phi_1(s_0+s_1+2\theta)$,
we get at convergences of learning, meaning $\nabla_{\phi_1}\mathcal{L}_\text{task} = 0$, the optimal parameter $\phi_1$ of $\mathcal{L}_\text{critic}$
\begin{equation}
\phi_1 = \frac{(s_0+2\theta-g)}{2(s_0+s_1+2\theta)}.
\end{equation}
See derivation in Appendix \ref{sec:app_details_mbml} for all the details. Remember, when the policy convergences - meaning that $\nabla_\theta \mathcal{L}_\text{critic} = 2\phi_1(s_0+s_1+2\theta) = 0$  - we get
\begin{equation}
    s_0+2\theta-g = 0 \Rightarrow \theta = \frac{g - s_0}{2}
\end{equation}
this is the optimal policy starting from $s_0$, the initial state

To conclude, our model-based bi-level meta-learning framework leads to parameter updates that lead to significantly different learned critics compared to supervised Q learning. In the presented toy example our meta-learning loop  recovers the optimal policy whereas standard supervised Q learning does not. 
Similar analysis for more general problems is left for future work. We empirically study the benefits of our approach for complex setups in section \ref{sec:experiments}.
\subsection{Illustration: Shape of $\mathcal{L}_\text{critic}$} \label{sec:what_learning}
To further analyze the representation of our learned $\mathcal{L}_\text{critic}$ we extend the toy example from above and represent the learned critic with a neural network $\mathcal{L}_\text{critic} = M(s_t,a_t,g,\phi)$ with parameters $\phi$. The policy is extended to two parameters $\pi(s) = [\theta_1,\theta_2]$. After learning $\mathcal{L}_\text{critic}$ with our algorithm, we compare the optimization landscape of $\mathcal{L}_\text{critic}$ with the landscape of the ground truth $Q_\text{true}$ function, which we compute analytically via Equation \ref{eq:Q_function}. We also compare $\mathcal{L}_\text{critic}$ with a neural network $Q_\text{sup}$ function learned in a supervised way as presented in section \ref{sec:related_work}. We train $\mathcal{L}_\text{critic}$ and $Q_\text{sup}$ on one goal from two initial states.

\begin{figure}
\centering
    \includegraphics[width=0.4\textwidth]{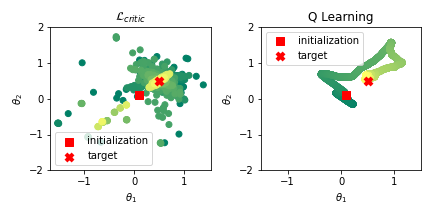}
    \vspace{-20pt}
    \caption{Sampling the parameters space during meta-training. The colors change from dark to light over the course of the training.}
    \label{fig:betas_training}
    \vspace{-15pt}
\end{figure}

Figure \ref{fig:point_mass_teaser} shows the optimization landscapes of $Q_\text{true}$, $Q_\text{sup}$ and our $\mathcal{L}_\text{critic}$. We plot the respective values over the policy parameter space $[\theta_1,\theta_2]=[x,y]$ where $x,y$ are represented by each grid in the plot. After rolling out the policy $\pi_{[x,y]}$ and observing the trajectory $\{s\}_{t=0}^T$, we plot $\sum_t^T Q_\text{true}(s_t,\pi_{[x,y]})$, $\sum_t^T Q_\text{sup}(s_t,\pi_{[x,y]})$ and $\sum_t^T \mathcal{L}_\text{critic}(s_t, \pi_{\theta=[x,y]})$ respectively. Darker colors correspond to lower cost and thus the optimal parameters of $\pi$ for the task. The arrows in the plot show the gradient fields of the policy gradients $\nabla_{[\theta_1,\theta_2]}Q_\text{true}$,$\nabla_{[\theta_1,\theta_2]}Q_\text{sup}$ and $\nabla_{[\theta_1,\theta_2]}\mathcal{L}_\text{critic}$ as a function of policy parameters. The arrows therefore show how the policy gradient behaves. From the plot it becomes visible that our $\mathcal{L}_\text{critic}$ very closely resembles the optimization landscape of the ground truth $Q_\text{true}$ function over the parameter space of $\mathbb{\boldsymbol{\theta}}$. This means that no matter how $\pi_\theta$ is initialized, the optimal values $\mathbb{\boldsymbol{\theta}}^*$ will be found when using our $\mathcal{L}_\text{critic}$. This essentially means that we are able to solve the reinforcement learning problem, namely to learn a policy $\pi$ that accomplishes task $g$, no matter where $\mathbb{\boldsymbol{\theta}}$ is being initialized. This is in contrast to $Q_\text{sup}$ which will over-fit to a neighborhood of the training goal and thus cannot be generally used to learn a policy that solves the task. In Figure \ref{fig:betas_training} we see how our method explores different parts of the policy parameter space $\mathbb{\boldsymbol{\theta}}$ before converging to the desired parameters $[\theta_1^*,\theta_2^*]$. This allows $\mathcal{L}_\text{critic}$ to learn about different parametrizations of $\pi$ and how to recover the optimal parameters for $g$.
\subsubsection{Generalization to new tasks}
Next we analyse the generalization capabilities of our $\mathcal{L}_\text{critic}$. In the previous section we have studied how $\mathcal{L}_\text{critic}$ is able to solve the RL task. Now we want to show how $\mathcal{L}_\text{critic}$ can generalize to different goals $g \in G$, after having been trained.
Since $\mathcal{L}_\text{critic}$ is learned in a goal-conditioned way and assuming $\mathcal{L}_\text{critic}$ learned the underlying learning mechanism to accomplish $g$, $\mathcal{L}_\text{critic}$ should be able to generalize to new $g \in G$ that have not been seen during training. Similarly, we can learn $Q_\text{sup}(s_t,a_t)$ in a goal-conditioned way by absorbing the goal information in $s \leftarrow s - g$. In Figure \ref{fig:point_mass_teaser} (bottom rows) again the optimization landscape of $Q_\text{true}$, $Q_\text{sup}$ and of our $\mathcal{L}_\text{critic}$ are shown for two different goals that were not seen during training. Our $\mathcal{L}_\text{critic}$ resembles the true $Q$ function's optimization landscape, even for new test goals. $Q_\text{sup}$ instead cannot generalize to new tasks. Our method learns a critic that allows for generalization to new and unseen situations - a crucial and very challenging open problem in RL. 
%
%
\begin{table*}
	\begin{center}
	\scalebox{0.7}{
		\begin{tabular}{c|c|c|c|c|c|c|c|c|c|c}
			Method & std=0.1 & std=0.2 & std=0.3 & std=0.4 & std=0.5 & std=0.6 & std=0.7 & std=0.8 & std=0.9 & std=1.0 \\
			\hline
			$\mathcal{L}_\text{critic}$ (ours) & \textbf{0.02(0.02)} & \textbf{0.04(0.03)} & \textbf{0.05(0.04)} & \textbf{0.07(0.05)} & \textbf{0.41(0.58)} & \textbf{1.65(2.63)} & \textbf{2.05(3.78)} & \textbf{1.01(1.92)} & \textbf{0.27(0.2)} & \textbf{0.95(0.8)} \\
			DDPG & 3.425(2.136) & 3.943(2.415) & 5.638(1.621) & 4.426(2.841) & 4.001(1.735) & 3.9(1.664) & 4.312(2.268) & 4.293(1.609) & 4.095(2.85) & 5.031(3.801) \\
			Meta Critic & 4.149(1.781) & 3.817(1.763) & 4.962(2.298) & 4.413(1.746) & 5.611(2.059) & 5.544(2.5) & 4.622(2.254) & 4.77(1.585) & 4.754(2.804) & 7.771(3.304) \\
		\end{tabular}}
	\end{center}
	\caption{Evaluation of $\mathcal{L}_\text{critic}$ on new tasks. 10 new goals where sampled at $\mathcal{N}(0,\text{std})$ around the training goals for $5$ seeds}\label{table:gen_tasks_kuka}
\end{table*}
\begin{table*}
	\begin{center}
	\scalebox{0.7}{
		\begin{tabular}{c|c|c|c|c|c|c|c|c|c|c}
			Method & Mass=0 & Mass=1 & Mass=2 & Mass=3 & Mass=4 & Mass=5 & Mass=6 & Mass=7 & Mass=8 & Mass=9 \\
			\hline
			$\mathcal{L}_\text{critic}$ (ours) & \textbf{0.066(0.058)} & \textbf{0.061(0.054)} & \textbf{0.06(0.053)} & \textbf{0.06(0.053)} & \textbf{0.06(0.053)} & \textbf{0.06(0.053)} & \textbf{0.06(0.053)} & \textbf{0.06(0.054)} & \textbf{0.061(0.054)} & \textbf{0.061(0.054)} \\
			DDPG & 3.688(1.817) & 4.276(1.295) & 5.132(0.595) & 3.494(2.585) & 2.141(1.489) & 1.882(0.841) & 4.503(3.101) & 3.919(2.408) & 3.96(2.321) & 2.327(2.229) \\
			Meta Critic & 1.495(0.105) & 2.696(0.865) & 3.31(1.198) & 2.879(0.621) & 3.274(0.409) & 4.267(1.215) & 4.312(0.375) & 3.205(1.314) & 2.674(0.912) & 2.238(0.704) \\
		\end{tabular}}
	\end{center}
	\caption{Generalization to different end-effector masses, training mass is at mass=3.12}\label{table:gen_kuka_mass}
\end{table*}
\begin{table*}
	\begin{center}
	\scalebox{0.7}{
		\begin{tabular}{c|c|c|c|c|c|c|c|c|c|c}
			Method & length=0.2 & length=0.4 & length=0.6 & length=0.8 & length=1.0 & length=1.2 & length=1.4 & length=1.6 & length=1.8 & length=2.0 \\
			\hline
			$\mathcal{L}_{critic}$ & \textbf{0.009(0.0)} & \textbf{0.009(0.0)} & \textbf{0.009(0.0)} & \textbf{0.009(0.0)} & \textbf{0.021(0.0)} & \textbf{0.021(0.0)} & \textbf{0.021(0.0)} & \textbf{0.021(0.0)} & \textbf{0.021(0.0)} & \textbf{0.021(0.0)} \\
			DDPG & 3.717(1.274) & 5.295(1.045) & 4.341(2.925) & 5.197(1.629) & 1.644(0.69) & 1.926(1.264) & 3.39(1.781) & 1.616(0.876) & 3.628(0.577) & 2.918(1.118) \\
			Meta Critic & 3.92(0.661) & 3.003(0.461) & 3.736(1.101) & 3.047(0.409) & 3.685(0.95) & 3.147(1.079) & 3.673(1.961) & 5.166(0.627) & 1.737(0.032) & 1.944(0.561) \\
		\end{tabular}}
	\end{center}
	\caption{Generalization to different end-effector link lengths, training value is at length=0}\label{table:gen_kuka_link}
\end{table*}

\section{Experiments}\label{sec:experiments}
We now present the experimental evaluations of our framework. We evaluate our framework on learning critics for a 2 degrees of freedom (DoF) planer reacher arm \cite{reacher} as well as a 7 DoF Kuka manipulator \cite{kuka} in simulation. In both cases, the task is to reach a desired goal position specified in joint space. The policy $\pi(s_t)=a_t$ is parametrized as a neural network where the input is the state $s=[\omega_{n=0\dots DoF},\dot{\omega}_{n=0\dots DoF}]$ composed of current joint positions and velocities and the output is the commanded joint torques. The policy neural network has two hidden layers with 64 neurons and tanh activation function. The learned critic $\mathcal{L}_\text{critic}$, also parametrized as a neural network, takes as an input $s$, the outputted $a$ from $\pi$ and the goal $g$. The neural network has two hidden layers with 400 neurons each and ELU activation function. We use the differentiable model presented in \cite{sutanto2020encoding} as our $f$. $\mathcal{L}_\text{task} = (s_T - g)^2$, where $g$ is the desired goal position defined in joint space. 

The experimental evaluation focuses of two aspects: (1) We analyse the capability of our model-based meta learning algorithm solve the RL task, this means during meta-train time a critic is learned that solves $g$. We show how using a model based approach increases sample efficiency and suffers less from variability in random seeds during meta-learning and (2) we show how our learned critic generalizes to tasks, dynamics and policy initializations that significantly differ from the meta-train time setting. 

We compare our method to two baselines:(1) \textbf{DDPG} \cite{lillicrap2015continuous}, where the $Q$ function is learned in a supervised way and the learned $Q$ function is used to learn the policy. Similarly we use our $\mathcal{L}_\text{critic}$ to learn the policy. (2) \textbf{Meta Critic} \cite{zhou2020online}, where also a meta critic is learned in a bi-level fashion and used to learn a policy. The meta critic is learned by evaluating the policy using a $Q$ function that was again learned in a supervised way. 
\subsection{Meta-train}
We start by analyzing how well each method can learn to solve the training tasks.
Each method is used to train a policy to reach four goals, and average results across 5 random seeds. Since the baselines are not inherently goal conditioned, we include the goal in the state description, thus the state becomes $s-g$ instead of $s$. 
\begin{figure}
    \centering
    \subfigure[Reacher]{\includegraphics[width=0.2\textwidth]{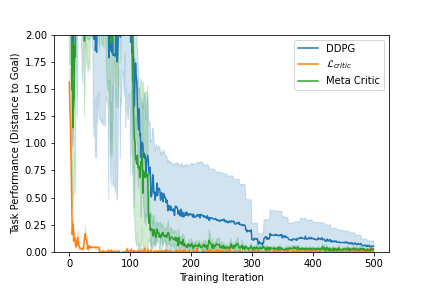}}\label{fig:reacher_training}
    \subfigure[Kuka]{\includegraphics[width=0.2\textwidth]{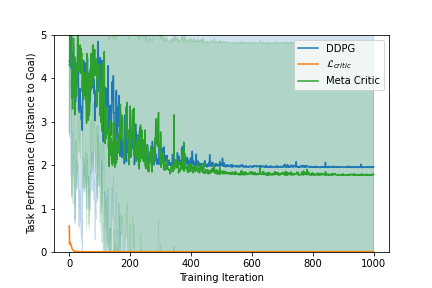}}\label{fig:kuka_training}
    \caption{Meta-training: Comparing $\mathcal{L}_\text{critic}$, DDPG and meta critic for both environments. Our method is more sample efficient.}\label{fig:meta_training}
    \vspace{-10pt}
\end{figure}
\vspace{-5pt}
\paragraph{2 DoF reacher and 7 DoF Kuka resultss}
Figure \ref{fig:meta_training} visualizes the training curves for both agents. Our model-based meta-learning approach solves the training task with fewer iterations and is thus more sample efficient compared to the baselines. We also observe that our method is the least sensitive to variations in random seeds. 
In summary, in both experiments our approach learns a critic that solves the training tasks more efficiently and reliably than the baselines.
%
\subsection{Meta-Test - Generalization}
We now turn to evaluating the generalization capabilities of our $\mathcal{L}_\text{critic}$ after meta-training. During meta-test $\mathcal{L}_\text{critic}$ is used to compute the policy gradient update
on a new policy $\pi$ on new settings. Similarly we use the critics learned with our baselines to compute the policy gradient for comparison. Specifically, we show two forms of generalization 1) generalization to new goals $g$ that have not been seen during training and 2) generalization to new (unknown) dynamics $\hat{f}$, that are different from the known dynamics model $f$ during meta-train time. 
For this set of experiments, we use the critics (q-functions) that were trained during meta-train for both the 2 DoF reacher arm and the 7 Dof Kuka arm.
\subsubsection{Generalizing to new tasks}
\begin{figure}
\hspace{-15pt}
    \includegraphics[width=0.5\textwidth]{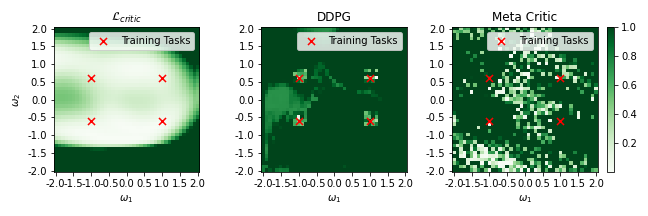}
    \caption{\small{reacher environment experiments on new tasks: generalization to new tasks over the full state space. The x and y axes represent the full state space. The red crosses are the training tasks. The colors show the error in accomplishing the new tasks. We observe that our $\mathcal{L}_\text{critic}$ generalizes far beyond the training tasks.}}
\label{fig:reacher_generalization}
\end{figure}

\paragraph{Results for 2 DoF reacher}
We test the generalization capabilities of our $\mathcal{L}_\text{critic}$ over the full state space of the reacher environment. Figure \ref{fig:reacher_generalization} shows the results, where each grid in the plot corresponds to a new goal state $s=[\omega_1,\omega_2]$.The crosses in the plots show the training goals that were used to train all critics. The colors in the plots represent the mean squared error in achieving the new goal, averaged over $5$ seeds. The error is measured as the distance to the desired goal $s=[\omega_1,\omega_2]$ at the end of the rollout of $\pi$, where $\pi$ was learned from scratch using $\mathcal{L}_\text{critic}$ or the critics learned with the baselines. We can see that our learned critic systematically generalizes to new tasks over the full state space. This is in contrast to the baselines that either overfit to the training tasks (DDPG) or do not reliably generalize to new tasks (Meta Critic) for new policies.
\paragraph{Results for 7 DoF Kuka}
In Table \ref{table:gen_tasks_kuka} we show the generalization capabilities of our $\mathcal{L}_\text{critic}$ to new tasks on the Kuka arm. We sample $10$ new tasks at a distance of $\mathcal{N}(0,\text{std})$ for $std=0.1\dots1.0$ around the training tasks. We use two of the four training tasks to sample around them. For each new task we learn 5 new policies, we use our $\mathcal{L}_\text{critic}$ as the policy gradient objective as described in algorithm \ref{algo:ml3_mbrl_test} and we do the same for the baselines. Using our $\mathcal{L}_\text{critic}$ on new unseen tasks allows for learning a policy that solves the task. This is in contrast to the baselines that do not perform well in this generalization experiment. We report the mean squared error between last state and goal over $10$ new goals, $5$ seeds and $5$ new policy initialization.
\subsubsection{Generalizing to different dynamics}
\begin{figure}
\hspace{-15pt}
    \includegraphics[width=0.5\textwidth]{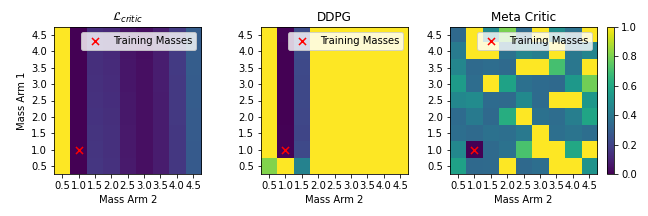}
    \caption{Generalization capabilities of our $\mathcal{L}_\text{critic}$ to new link masses. We evaluate the generalization capabilities over different combination of masses of the links. The red cross shows the combination of masses that was used during training.}
    \label{fig:reacher_mass_gen}
    \vspace{-15pt}
\end{figure}

\begin{figure}
\hspace{-15pt}
    \includegraphics[width=0.5\textwidth]{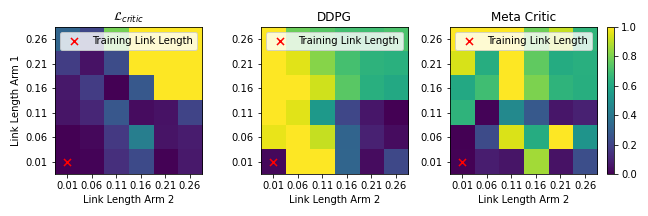}
    \caption{Generalization capabilities of our $\mathcal{L}_\text{critic}$ to new link length. The red cross shows the combination of masses that was used during training.}
    \label{fig:reacher_link_gen}
    \vspace{-15pt}
\end{figure}

\paragraph{Results for 2 DoF reacher}
We evaluate how well our $\mathcal{L}_\text{critic}$ generalizes to learning new policies for agents with new link masses. In Figure \ref{fig:reacher_mass_gen} the errors are shown to accomplish the training task under new dynamics. Each point in the figure represents a new combination of masses of the links. The red cross shows the training combination. Our learned critic generalizes reliable to heavier masses, which is not the case for the baselines.
We also evaluated the generalization performance under changing link lengths. In Figure \ref{fig:reacher_link_gen} again each point represents a different combination of link lengths, the red cross represents the training combination. Varying the length of the links significantly alters the dynamics. Also in this scenario our $\mathcal{L}_\text{critic}$ generalizes better to new lengths when compared to the baselines.
\vspace{-10pt}
\paragraph{Results for 7 Dof Kuka}
In this set of experiment we evaluate how well our learned loss can generalize to new dynamics on the 7 DoF kuka arm. We vary the mass and the link length at the end-effector. This is comparable to varying the weight and shape of an object in the robot's hand, for example when wanting to place a cup instead of a bottle. In Table \ref{table:gen_kuka_mass} the results for the changes in mass, in Table \ref{table:gen_kuka_link} the results of the changes in link length are reported and compared to the baselines. Our $\mathcal{L}_\text{critic}$ generalizes also to new mass and length values at the end-effector, which is not true for the baselines. Again we report the mean squared error between last state and goal over $10$ new goals, $5$ seeds and $5$ new policy initialization. To make the comparison fairer, for our baselines, we selected seeds that solved the task during training. 
\vspace{-10pt}
\section{Conclusion and Discussion}
Generalization in RL is an important and fairly new research direction \cite{kirk2021survey} but crucial for deploying RL in the real world. In this work we present a model-based meta-learning framework to learn a critic that is used for policy optimization in RL. In contrast to previous work, our approach meta-learns the critic by directly following the learning progress of the policy leveraging the model during meta-training. During meta-test time the learned critic can be used to learn new policies for new tasks and dynamics. The results show the generalization capabalities of our learned critic. Notably our learned critic can be used at meta test time in a model free way while still being able to adapt to changes in the dynamics. This could make it a great contender for sim-to-real transfer. In the future we want to explore the use of simpler models during meta-train time and include a model-learning component during training.


\bibliography{main}
\bibliographystyle{icml2022}

\newpage
\appendix
\onecolumn

\section{Analytical Analysis Details}

We define the following learned critic $\mathcal{L_\text{critic}}$, model $f$, policy $\pi$, sparse reward $r$ and learned $Q$ function.
\begin{equation}
    \begin{split}
        \mathcal{L_\text{critic}}(s_t,a_t) = Q(s_,a_t) = (s_t + a_t)^2\phi_1\\
        f(s_t,a_t) = s_t+a_t = s_{t+1}\\
        a_t = \pi_\theta(s_t) = \theta \\
        r_t(s_t,a_t) =\begin{cases}
      0 & \text{if }t<T \\
      (s_t-g)^2 & \text{if }t=T\\
    \end{cases}     
    \end{split}
\end{equation}
Where $g$ is the goal. We consider a time horizon of $T=2$, following the example above at $t=T=2$ the goal will be $g = s_0 + 2\theta$ thus the optimal policy to reach a desired goal should be $\pi(s) = \theta = \frac{g - s_0}{2} $. We use the policy $\pi_\theta(s) = \theta$ to collect the roll-out $\tau = \{s_0,a_0, r_0, s_1, a_1, r_1, s_2, r_2\}$ using model $f$. The goal of this simple example is to see how a global optimum of the learning loops would look like if it existed. To do this, we train the critics to convergence and use the optimal parameters of the critics to train the policy to convergence. 

\subsection{Q Learning Analysis Details}\label{sec:illapp_q_fitting}
Let's first consider the case where we learn $Q$ from data, as presented in section \ref{sec:sup_learning_q}. For this we define the $Q$ function with parameter $\phi_1$:
\begin{equation}
\begin{split}
    Q(s_t,a_t) = (s_t+ a_t)^2 \phi_1
\end{split}
\end{equation}

The objective of $Q$ learning is defined in Equation \ref{eq:q_error_function_sup} and we compute it using the current data rollout $\tau$. With that the supervised $Q$ learning loss $\delta$ is then 
\begin{align}
\delta &= \sum_{t=0}^{1} (Q_t(s_t,a_t) - (r_t + \gamma Q_{t+1}(s_{t+1},a_{t+1}^2)))^2 \\
&= ((s_0 + a_0)^2\phi_1 - (r_0 + (s_1+ a_1)^2 \phi_1))^2+ 
((s_1+ a_1)^2\phi_1 - (r_1 + r_2))^2 \\
&= ((s_0 + a_0)^2\phi_1 - (s_1+ a_1)^2\phi_1))^2+ 
((s_1+ a_1)^2\phi_1 - r_2))^2
\end{align}
since $r_0=r_1=0$ and we assume $\gamma=1$. Using the knowledge that all actions are the same $a_0 = a_1$, we can remove the time index on actions and simplify the error function to 
\begin{align}
    \delta &= ((s_0 + a)^2\phi_1 - (s_1+ a)^2\phi_1))^2+ 
((s_1+ a)^2\phi_1 - r_2))^2
\end{align}

Taking the gradient of $\delta$ wrt. $\phi_1$ results in:
\begin{align}\label{eq:simple_q_derivation}
    \nabla_{\phi_1} \delta &= 2\phi_1[(s_0+a)^2 - (s_1+a)^2][(s_0+a)^2 - (s_1+a)^2] + 2[\phi_1(s_1+a)^2 - r_2](s_1+a)^2 \\
    &=2\phi_1[((s_0+a)^2-(s_1+a)^2)^2 + (s_1+a)^4] - 2(s_1+a)^2r_2
\end{align}
to find the optimal Q function parameters we set the gradient to 0, and solve for $\phi_1$:
\begin{align}
    \nabla_{\phi_1} \delta &= 0\\
    2\phi_1[((s_0+a)^2-(s_1+a)^2)^2 + (s_1+a)^4] - 2(s_1+a)^2r_2 &= 0 \\ 
     \phi_1 = \frac{(s_1+a)^2r_2}{[((s_0+a)^2-(s_1+a)^2)^2 + (s_1+a)^4]}
\end{align}

Now we compute the gradient for policy parameters $\theta$ with respect to the predictions of the learned $Q$ function: 
\begin{align}
    \nabla_\theta \sum_{t=0}^{T=1} Q(s_t, a_t) =  \nabla_\theta ((s_0 + \theta)^2 \phi_1 + (s_1+\theta)^2 \phi_1 + r_2)
 = 2\phi_1[(s_0+\theta)+(s_1+\theta)]
 \end{align}
where we have expanded the actions in the Q function predictions with the policy parametrization - such that we can compute the analytical gradients wrt to $\theta$.

At convergence, the gradient $\nabla_\theta \sum_{t=0}^{T=1} Q(s_t, a_t) = 2\phi_1[(s_0+\theta)^2+(s_1+\theta)^2]=0$. To analyse the convergence, we use the optimal $Q$ function parameter $\phi_1$ that we computed before:
\begin{align}\label{app_eq:convergence_q}
     \phi_1&=\frac{(s_1+a)^2r_2}{[((s_0+a)^2-(s_1+a)^2)^2 + (s_1+a)^4]}\\
     &\Rightarrow 2\phi_1[(s_0+\theta)+(s_1+\theta)]= 2\frac{(s_1+a)^2r_2[(s_0+\theta)+(s_1+\theta)]}{[((s_0+a)^2-(s_1+a)^2)^2 + (s_1+a)^4]}=0 \\
     &\Rightarrow (s_1+a)^2r_2[(s_0+\theta)+(s_1+\theta)]=0
\end{align}
This training procedure may converge, however we do not necessarily recover the optimal policy. Solving for $\theta$ we find that the loop converges when:
\begin{align}
    [(s_0+\theta)+(s_1+\theta)]=0 \Rightarrow \theta = \frac{s_0+s_1}{-2}
\end{align}
This is not the optimal policy.

We analyse this further for the case where we have access to the model $f(s_t,a_t)=s_t+\theta=s_{t+1}$ during $Q$ learning. This means the reward $r_2=(s_0+2\theta-g)^2$ now is also a function of $\theta$. The derivation of above stays the same and $\theta = \frac{s_0+s_1}{-2}$ is still a solution this loop converges to. However the loop now could also converges when $r_2=0 \rightarrow (s_0+2\theta-g)^2=0 \rightarrow  \theta=\frac{-s_0+g}{2}$. 
While the optimal policy could be recovered in this case, learning still does not necessarily converge to that. 

\begin{figure}
\centering
    \includegraphics[width=0.5\textwidth]{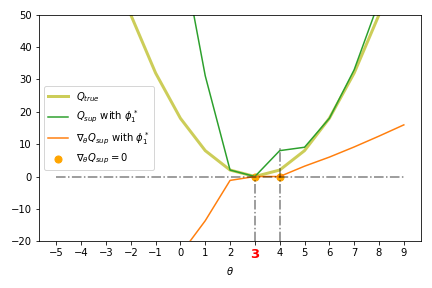}
    \caption{The learned Q is shown after learning converged and the optimal parameters $\phi_1$ have been computed. We also show the gradient of $Q$ wrt $\theta$ $\nabla_\theta Q$. From the graph it becomes visible that the $Q$ learned in a supervised way has two minima, a global minimum where the optimal policy would be recovered at $\theta=3$ and a second local minima the optimization could get stuck in at $\theta=4$. This shows that the learning procedure does not necessarily recover the optimal policy.}
    \label{fig:ill_q_sup}
\end{figure}

In Figure \ref{fig:ill_q_sup} we show the graphs of $Q$ and $\nabla_\theta Q$ after supervised learning of $Q$ converged. We also show the true $Q$ function, that we can compute given the simplicity of the example as $Q_\text{true}=r_0+r_1+r_2$. We show a practical example where $s_0=-6$ and $g=0$. We can see that this optimization has two points where $\nabla_\theta Q=0$ and thus two minima. The global optimum where $\theta=3$ is the desired one and recovers the optimal policy. However, without access to the model $f$, this optimum will only be found if $r2=0$ to begin with, which means that we initialize the policy with the optimal one. If we have access to $f$, as discussed above, the optimal policy can be found but is not the only point of convergence of the learning loop.

\subsection{Model Based Meta-Learning the Critic Analysis Details}\label{sec:app_details_mbml}
We analyze the learning procedure of our method. We first update the parameters $\theta$ using our $\mathcal{L}_\text{critic}$ (in the inner loop). To do so, we compute 
\begin{align}
    \nabla_\theta \sum_{t=0}^{T=1}\mathcal{L}_\text{critic} = \nabla_\theta [(s_0+\theta)^2\phi_1 + (s_1+\theta)^2\phi_1] = 2\phi_1(s_0+s_1+2\theta)
\end{align}
The updated $\theta$ parameters are given by 
\begin{align}
    \theta_\text{new} = \theta_\text{old} - \alpha \nabla_\theta \sum_{t=0}^{T=2}\mathcal{L}_\text{critic} \Rightarrow \theta_\text{old} - \phi_1(s_0+s_1+2\theta)
\end{align}
 where we assume $\alpha=\frac{1}{2}$.
As in Algorithm \ref{algo:ml3_mbrl} we use $\theta_{new}$ to compute $\mathcal{L}_\text{task}$ defined per Equation \ref{eq:cumulative_reward}:
\begin{equation}
    \mathcal{L}_\text{task} = \sum_t r_t = r_0 + r_1 + r_2 = r_2 = (s_2-g)^2
\end{equation}
expanding all terms, meaning $s_2 = s_1+\theta_\text{new}$, $s_1=s_0+\theta_\text{new}$, $\theta_\text{new} = \theta-\phi_1(s_0+s_1+2\theta)$ we get: 
\begin{align}
s_2 & = s_0+2\theta-2\phi_1(s_0+s_1+2\theta)\\
    & \mathcal{L}_\text{task} = (s_2 -g)^2 = (s_0+2\theta-2\phi_1(s_0+s_1+2\theta) -g)^2 
\end{align}

Now we want to find the optimal parameters $\phi$ by taking the gradient $\nabla_{\phi_1} \mathcal{L}_\text{task}$ and we get 
\begin{align}
\nabla_{\phi_1} \mathcal{L}_\text{task} & = 2(s_2-g)[-2(s_0+s_1+2\theta)] \\
& =(s_0+2\theta-2\phi_1(s_0+s_1+2\theta) -g)[-2(s_0+s_1+2\theta)]\\
\end{align}
and at convergences of learning, meaning $\nabla_{\phi_1}\mathcal{L}_\text{task} = 0$ we get 
\begin{align}
&(s_0+2\theta-2\phi_1(s_0+s_1+2\theta) -g)[-2(s_0+s_1+2\theta)]=0\\
&\Rightarrow 4\phi_1[(s_0+s_1+2\theta)(s_0+s_1+2\theta)]=2[(s_0+2\theta-g)(s_0+s_1+2\theta)]\\
&\Rightarrow \phi_1 = \frac{[(s_0+2\theta-g)(s_0+s_1+2\theta)]}{2[(s_0+s_1+2\theta)(s_0+s_1+2\theta)]} = \frac{(s_0+2\theta-g)}{2(s_0+s_1+2\theta)}
\end{align}
thus, recover the optimal parameter $\phi_1$ of $\mathcal{L}_\text{critic}$ at convergence.

Remember, when the policy convergences - meaning that $\nabla_\theta \mathcal{L}_\text{critic} = 2\phi_1(s_0+s_1+2\theta) = 0$. Using the previously computed optimal $\phi_1$ we get:
\begin{align}
    & 2\phi_1(s_0+s_1+2\theta) = 0 \\
    & \Rightarrow 2\frac{(s_0+2\theta-g)}{2(s_0+s_1+2\theta)}(s_0+s_1+2\theta)=0 \\
    &  \Rightarrow s_0+2\theta-g = 0 \Rightarrow \theta = \frac{-s_0+g}{2}
\end{align}
this is the optimal policy starting from $s_0$, the initial state.

\begin{figure}
\centering
    \includegraphics[width=0.5\textwidth]{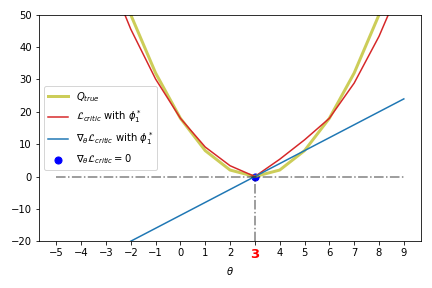}
    \caption{Our $\mathcal{L}_\text{critic}$ is plotted after learning converged and the optimal parameters $\phi_1^*$ have been found. We also plot $\nabla_\theta \mathcal{L}_\text{critic}$. From the graph it becomes visible that the learned $\mathcal{L}_\text{critic}$ has only one global minimum where the optimal policy is recovered at $\theta=3$. This shows that the learning procedure at convergence recovers the optimal policy.}
    \label{fig:ill_critic}
\end{figure}

In Figure \ref{fig:ill_critic} we show the graphs of $\mathcal{L}_\text{critic}$ and $\nabla_\theta \mathcal{L}_\text{critic}$ after learning of $\mathcal{L}_\text{critic}$ converged as a function of $\theta$. We also show the true $Q$ function, that we can compute given the simplicity of the example as $Q_\text{true}=r_0+r_1+r_2$. We preset a practical example where $s_0=-6$ and $g=0$. We can see that this optimization has only one point where $\nabla_\theta \mathcal{L}_\text{critic}=0$ and thus one global minimum at optimal policy parameters $\theta=3$. We also see that our $\mathcal{L}_\text{critic}$ resembles in shape $Q_\text{true}$.

\subsection{Meta Critic Analysis}\label{sec:ill_meta_critic}
In the meta critic learning algorithm presented in \cite{zhou2020online}, two critics are learned, a regular critic $Q(s_t,a_t)$ and a meta critic $Q_\text{meta}(s_t,a_t)$. Both are used to update the policy $\pi$. $Q(s_t,a_t)$ is learned with the supervised learning update rule, and thus follows the same derivation as in \ref{sec:analysis_simple_q}. We define:
\begin{align}
    Q(s_t,a_t) = (s_t + a)^2\phi_1 \text{ as previously} \\
    Q_\text{meta} = (s_t+ a)^2\beta_1
\end{align}
In this algorithm the policy is updated using both critics $\theta_\text{new} = \theta - \nabla_\theta \sum_{t=0}^{T=1}(Q(s_t,a_t) + Q_\text{meta}(s_t,a_t)) = \theta - 2(\phi_1+\beta_1)((s_0+a)^2+(s_1+a)^2)$. At convergence, when $\nabla_\theta (Q(s_t,a_t) + Q_\text{meta}(s_t,a_t))=0 \Rightarrow \phi_1 = - \beta_1$.

To update the meta critic, the updated policy $\pi_\text{new} = \theta - 2(\phi_1+\beta_1)((s_0+a)^2+(s_1+a)^2)$ is used and evaluated using $Q(s_t,a_t)$. 
If we use $\theta_\text{new}$ the learning signal for $Q_\text{meta}$ is
\begin{align}
    l = \sum_{t=0}^1 (s_t + [\theta - 2(\phi_1+\beta_1)((s_0+a)^2+(s_1+a)^2)])^2\phi_1
\end{align}
thus
\begin{align}
  \sum_t \nabla_{\beta_1} l &= 2((s_t + [\theta - 2(\phi_1+\beta_1)((s_0+a)^2+(s_1+a)^2)]))[((s_0+a)^2+(s_1+a)^2)] \\
  &=2[((s_0+a)^2+(s_1+a)^2)] \sum_t ((s_t + [\theta - 2(\phi_1+\beta_1)((s_0+a)^2+(s_1+a)^2)])) \\
  &=2[((s_0+a)^2+(s_1+a)^2)] \sum_t[(s_t)] +2[\theta - 2(\phi_1+\beta_1)((s_0+a)^2+(s_1+a)^2)]\\
\end{align}
Setting $\nabla_{\beta_1} l=0$
\begin{align}
s_0+s_1 + 2\theta((s_0+a)^2+(s_1+a)^2) - 4\phi_1((s_0+a)^2+(s_1+a)^2) = 4\beta_1((s_0+a)^2+(s_1+a)^2) \\
\Rightarrow \beta_1 = \frac{s_0+s_1}{4[(s_0+a)^2+(s_1+a)^2]} + \frac{\theta}{2} - \phi_1
\end{align}.

The critic $Q(s_t,a_t)$ is updated the same way as in Section \ref{sec:analysis_simple_q}. Since the policy converges when $\phi_1=-\beta_1$ we get:
\begin{align}
    - \phi_1 = \frac{s_0+s_1}{4[(s_0+a)^2+(s_1+a)^2]} + \frac{\theta}{2} - \phi_1 \\
    \Rightarrow 0 = \frac{s_0+s_1}{4[(s_0+a)^2+(s_1+a)^2]} + \frac{\theta}{2}\\
    \Rightarrow \theta = -\frac{s_0+s_1}{2[(s_0+a)^2+(s_1+a)^2]}
\end{align}
This does not recover the optimal policy.
\end{document}